\documentclass[conference]{IEEEtran}
\IEEEoverridecommandlockouts
\usepackage{cite}
\usepackage{amsmath,amssymb,amsfonts}
\usepackage{algorithmic}
\usepackage{graphicx}
\usepackage{textcomp}
\usepackage{xcolor}
\usepackage{colortbl}
\usepackage{subfig}
\def\BibTeX{{\rm B\kern-.05em{\sc i\kern-.025em b}\kern-.08em
    T\kern-.1667em\lower.7ex\hbox{E}\kern-.125emX}}
\begin{document}

\title{\textbf{E-ReCON:} An Energy- and Resource-Efficient Precision-Configurable Sparse nvCIM Macro for Conventional and Spiking Neural Edge Inference}

\author{
\IEEEauthorblockN{Ankit Kumar Tenwar, Mukul Lokhande, Santosh Kumar Vishvakarma}
\IEEEauthorblockA{\textit{Dept. of Electrical Engineering} \\
\textit{Indian Institute of Technology Indore} \\
Simrol 453552, Indore, India \\
\{phd2301202005, phd2201102020, skvishvakarma\}@iiti.ac.in}\\
\thanks{This work was supported partially by the Dept of Science and Technology (DST), Govt of India, for the INSPIRE PhD fellowship, and MeitY/SMDP-C2S for ASIC design tools.}}

\maketitle

\begin{abstract}
This work presents E-ReCON, a 16 Kb energy- and resource-efficient digital compute-in-memory (DCIM) macro based on a compact 3T1R ReRAM bitcell for edge-AI inference. The proposed bitcell occupies only 0.85 $\mu$m\textsuperscript{2} and supports reliable AND-based in-memory multiplication for both conventional convolutional neural network (CNN) and spiking neural network (SNN) workloads. To reduce accumulation overhead, a novel interleaved 10T/28T adder tree is introduced, reducing transistor count and power consumption by 37\% and 28\%, respectively, compared to a conventional 28T RCA-based design. Implemented in 65 nm CMOS at 1.2 V, the proposed macro achieves a minimum latency of 0.48 ns, throughput of 2.31-3.1 TOPS, and energy efficiency of up to 419 TOPS/W. When evaluated on LeNet-5, AlexNet, and CNN-8 models, the macro achieves 97.81\%, 93.23\%, and 96.51\% accuracy on MNIST/A-Z, CIFAR-10, and SVHN datasets, respectively. In addition, 40\% pruning preserves nearly 99.8\% of the original accuracy while reducing MAC operations and computation cycles. For SNN-oriented workloads, the proposed AND-type bitcell efficiently supports spike-weight multiplication with low switching activity, where the 2A2W configuration achieves accuracy close to the FP32 baseline across VGG-8, VGG-16, and ResNet-18 networks on CIFAR-10, CIFAR-100, and ImageNet-1K datasets. Compared to prior ADC-based ReRAM-CIM designs, the proposed architecture improves latency and energy efficiency by nearly 30-40\% while maintaining robust operation under full PVT and ReRAM variability. Overall, E-ReCON provides a scalable, low-latency, and energy-efficient nvCIM platform for next-generation edge-AI, IoT, biomedical sensing, and neuromorphic applications.
\end{abstract}

\begin{IEEEkeywords}
Non-volatile Memory, multiply-accumulate, Edge Neural Computing, Energy-efficient, ReRAM. 
\end{IEEEkeywords}

\section{Introduction}

Artificial Intelligence (AI) workloads are increasingly being deployed on edge devices for applications such as image classification, speech recognition, biomedical sensing, and autonomous monitoring. These workloads rely heavily on convolutional neural networks (CNNs) and other neural models that require billions of multiply-and-accumulate (MAC) operations. However, conventional Von-Neumann architectures suffer from significant latency and energy overhead due to frequent data movement between memory and processing units, making them inefficient for resource-constrained edge platforms \cite{Flex-PE}. To address this challenge, compute-in-memory (CIM) architectures have emerged as a promising alternative by performing computation directly inside the memory array, thereby reducing data transfer, improving throughput, and enhancing energy efficiency \cite{huo2022computing, tenwar-vdat, pratham-apccas}.

Among various non-volatile memory (NVM) technologies, such as MRAM, FERAM, PCRAM, and ReRAM, ReRAM has attracted considerable attention for CIM systems because of its low operating voltage, fast switching speed, compact footprint, and CMOS compatibility \cite{milozzi2024memristive, REFLEX-PIM, mu20241}. Furthermore, the non-volatility of ReRAM enables low standby power and reduces the need for costly data transfers between external NVM, DRAM, and SRAM-based compute units. By directly storing neural weights inside the ReRAM array, off-chip memory accesses can be significantly reduced for edge-AI devices with strict energy and latency constraints.

Typical ReRAM-based CIM architectures rely on analog current accumulation at the bitline to perform MAC operations. Although such approaches provide high density, they often suffer from low signal-to-noise ratio (SNR), process-voltage-temperature (PVT) variations, ADC overhead, and limited precision \cite{kim2021colonnade, FERMI,  sharma202264}. Current-domain CIM offers high parallelism but suffers from large read currents and degraded energy efficiency at larger array sizes \cite{wang2021efficient}. Charge-domain CIM reduces static current through capacitive accumulation or 2T2R structures, but introduces area overhead and additional peripheral circuitry \cite{ezzadeen2021low}. Time-domain CIM improves robustness against current summation errors, but still requires high-resolution ADCs, increasing area and latency \cite{XAC-MAC-BPD-tnano23}.

Digital SRAM-based DCIM architectures overcome many of the precision and noise limitations of analog approaches by providing deterministic computation \cite{12T-NOR-ISSCC'22, 2Tmult-Date25}. However, analog ReRAM-CIM architectures still rely heavily on ADCs or time-multiplexed conversion schemes, which reduce throughput and increase peripheral overhead \cite{yu2021compute}. To address these limitations, ADC-less digital ReRAM-based CIM macros are becoming increasingly attractive. In addition to conventional CNN workloads, the proposed AND-based 3T1R nvCIM bitcell is also suitable for spiking neural network (SNN) inference. Since spike-based computation relies on binary activations and event-driven accumulation, the proposed bitcell can directly support spike-weight multiplication using the same AND-based operation. Similar to prior AND-based SNN-oriented CIM works \cite{spike_and}, the proposed architecture enables low-power spike-domain accumulation with reduced switching activity and memory access overhead.

This work proposes a 16 Kb ReRAM-based digital compute-in-memory (DCIM) macro in 65 nm CMOS technology for efficient edge-AI inference. The proposed architecture eliminates ADC/DAC overhead through a compact 3T1R digital ReRAM bitcell and an interleaved 10T/28T adder tree for low-area accumulation. The major contributions of this work are summarised as follows:

\begin{itemize}
    \item \textbf{Proposed DCIM architecture:} A fully digital ReRAM-CIM architecture that eliminates iterative tuning and ADC/DAC requirements for deterministic low-latency operation.

    \item \textbf{Support for conventional and spiking inference:} The proposed 3T1R bitcell supports both CNN and SNN workloads through efficient AND-based in-memory multiplication.
    
    \item \textbf{A dual-purpose 3T1R AND-type ReRAM bitcell:} The proposed bitcell improves computational density, minimises peripheral overhead, and provides a smaller footprint than advanced multi-device structures such as 9T4R \cite{9T4R-TCASI'25}.
        
    \item \textbf{An area-efficient interleaved adder tree:} A resource-efficient interleaved 10T/28T adder tree reduces transistor count by nearly 37\% and lowers power consumption compared to a conventional 28T RCA-based structure.
\end{itemize}

The remainder of this paper is organised as follows. Section II presents the proposed bitcell, adder tree, and DCIM macro. Section III describes the evaluation methodology and experimental results, while Section IV concludes the paper.

\begin{table}[!t]
\caption{Truth table of the proposed 3T1R ReRAM bitcell showing AND-based in-memory multiplication for different input activations and stored weight states.}
\resizebox{\columnwidth}{!}{%
\label{tab:mac-compute}
\centering
\begin{tabular}{c|c|c|c|c|c|c}
\hline
\multicolumn{3}{c|}{\textbf{Input}} & \multicolumn{2}{c|}{\textbf{Weight}} & \multicolumn{2}{c}{\textbf{Output}} \\ \hline
\textbf{Value} & \textbf{BL} & \textbf{SL} & \textbf{Value} & \textbf{ReRAM State} & \textbf{Value} & \textbf{MULT (AND)} \\ \hline
1    & V\textsubscript{LOW}  & V\textsubscript{HIGH} & 1    & LRS & 1    & VDD \\ \hline
1    & V\textsubscript{LOW}  & V\textsubscript{HIGH} & -1 (0)  & HRS & -1 (0)  & GND \\ \hline
-1 (0)  & V\textsubscript{HIGH} & V\textsubscript{LOW}  & 1    & LRS & -1 (0)  & GND \\ \hline
-1 (0)  & V\textsubscript{HIGH} & V\textsubscript{LOW}  & -1 (0)  & HRS & -1 (0)   & GND \\ \hline
\end{tabular}}
\end{table}

\begin{figure}[!t]
    \centering
    \subfloat[]{\includegraphics[width=0.4\columnwidth]{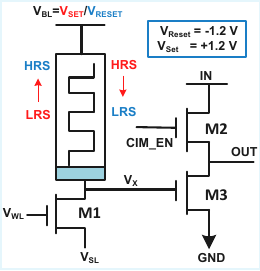}
        \label{fig:LRS}}
    \centering
    \subfloat[]{\includegraphics[width=0.54\columnwidth]{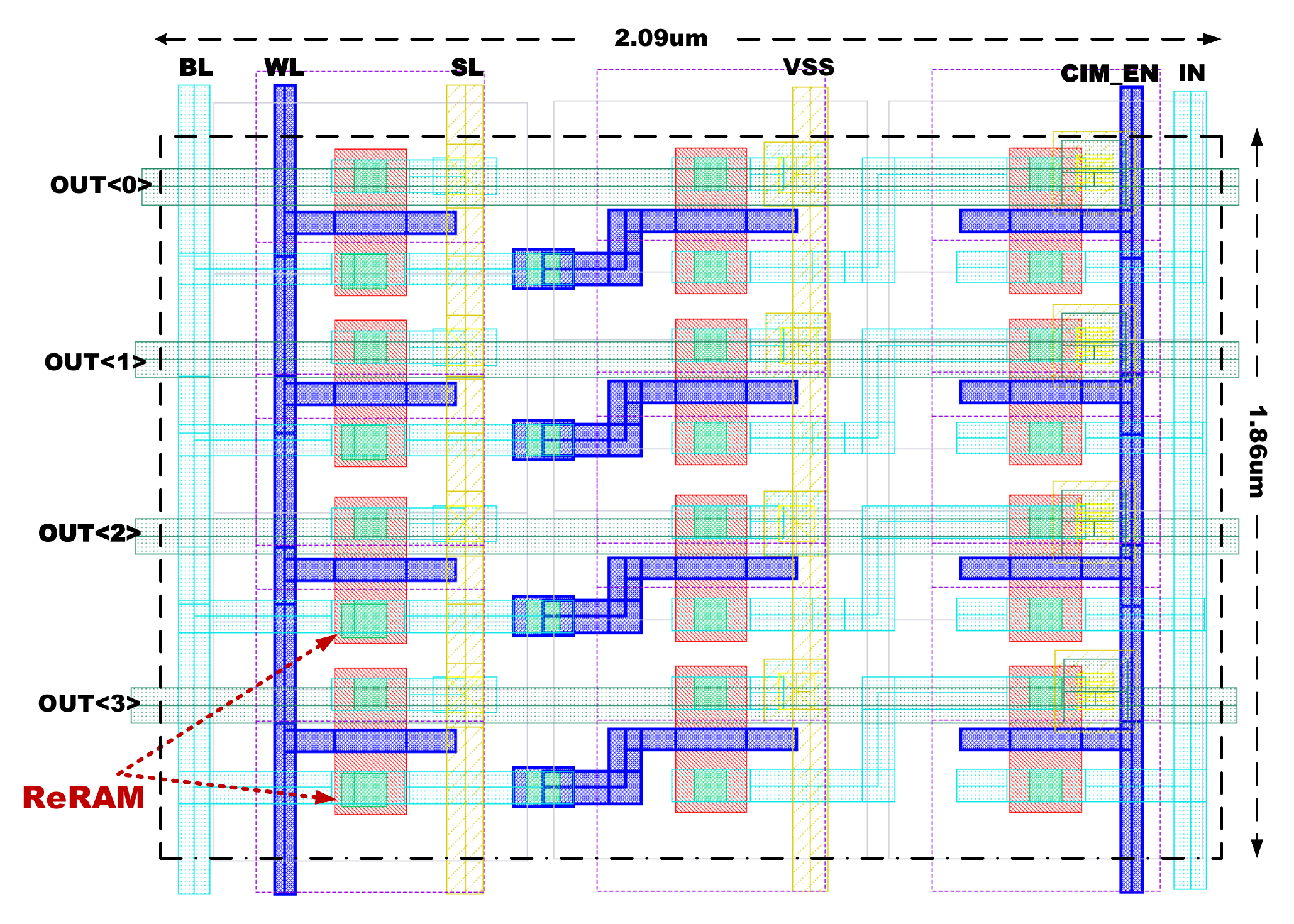}
    \label{fig:HRS}}
    \caption{(a) Schematic of the proposed CIM-enabled 3T1R ReRAM bitcell implementing resistive AND-based multiplication, (b) Proposed digital ReRAM unit (1 × 4) layout.}
\end{figure}

\begin{figure}[!t]
    \centering
    \subfloat[]{\includegraphics[width=0.48\columnwidth, height=5.5cm]{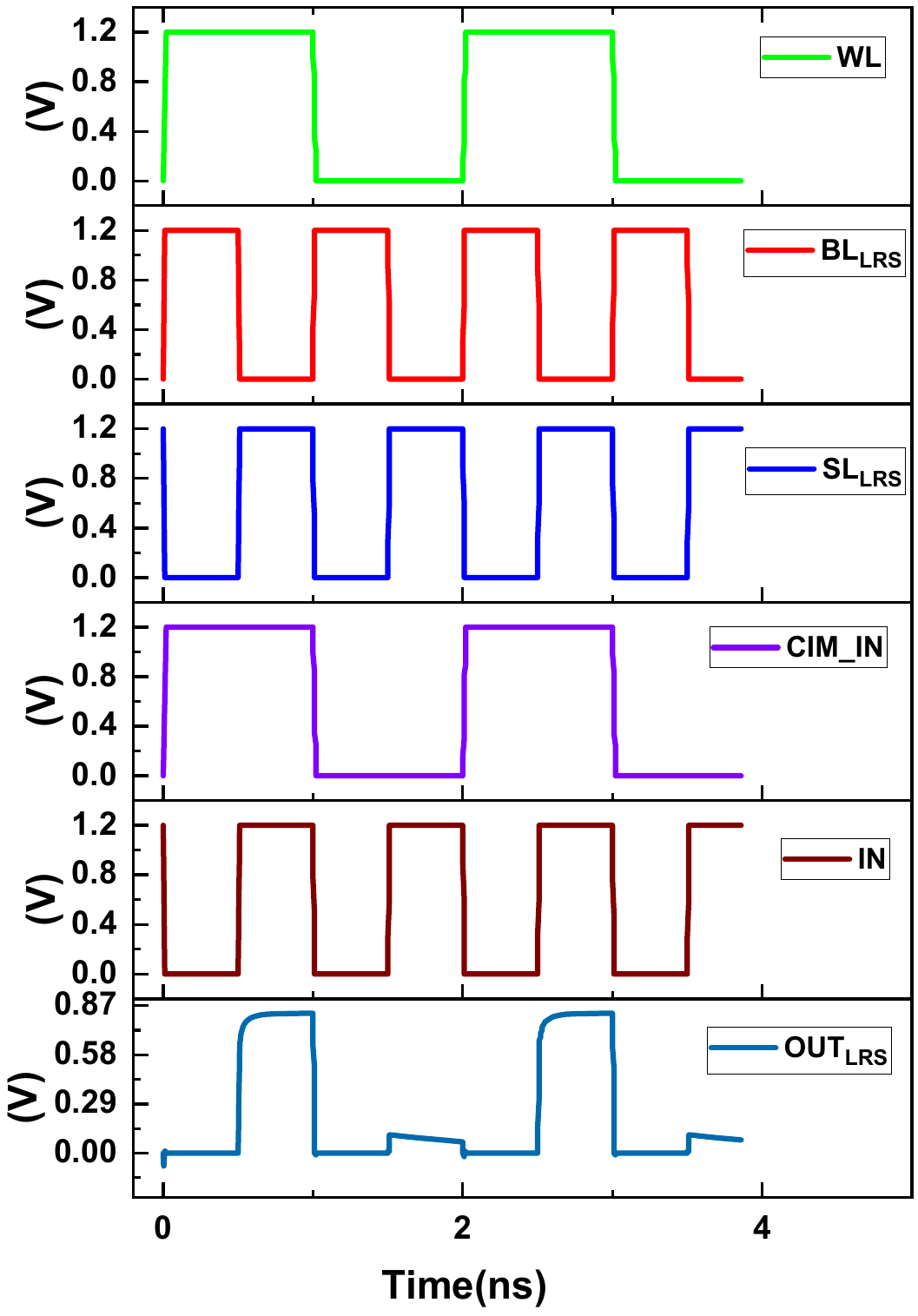}
        \label{fig:LRS}}
    \centering
    \subfloat[]{\includegraphics[width=0.44\columnwidth]{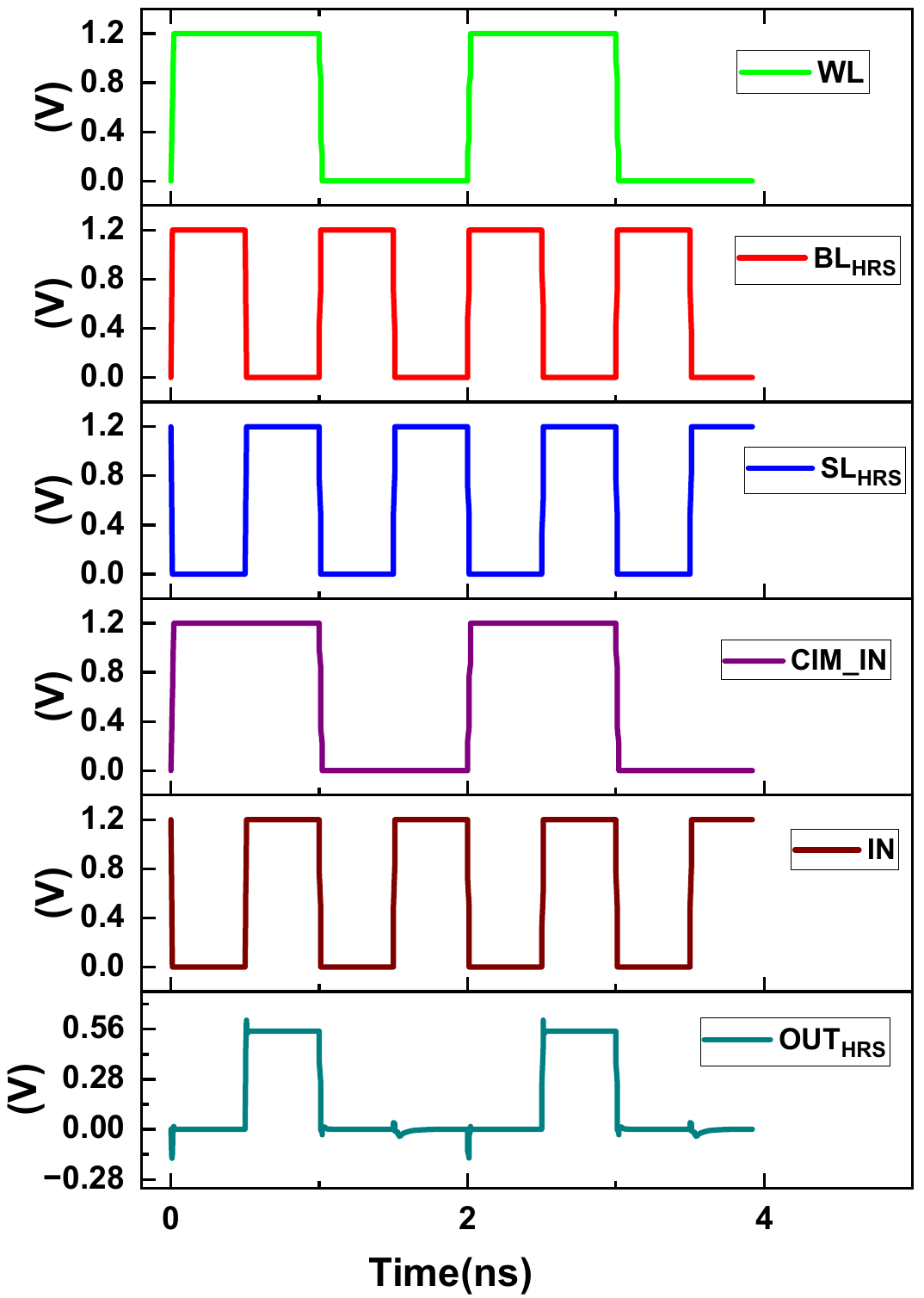}
    \label{fig:HRS}}
    \caption{Transient simulation waveforms of the proposed 3T1R bitcell during MAC operation for (a) HRS-stored weight (‘0’) and (b) LRS-stored weight (‘1’).}
\end{figure}

\begin{figure}[!t]
    \centering
    \subfloat[]{\includegraphics[width=0.46\columnwidth, height=37.5mm]{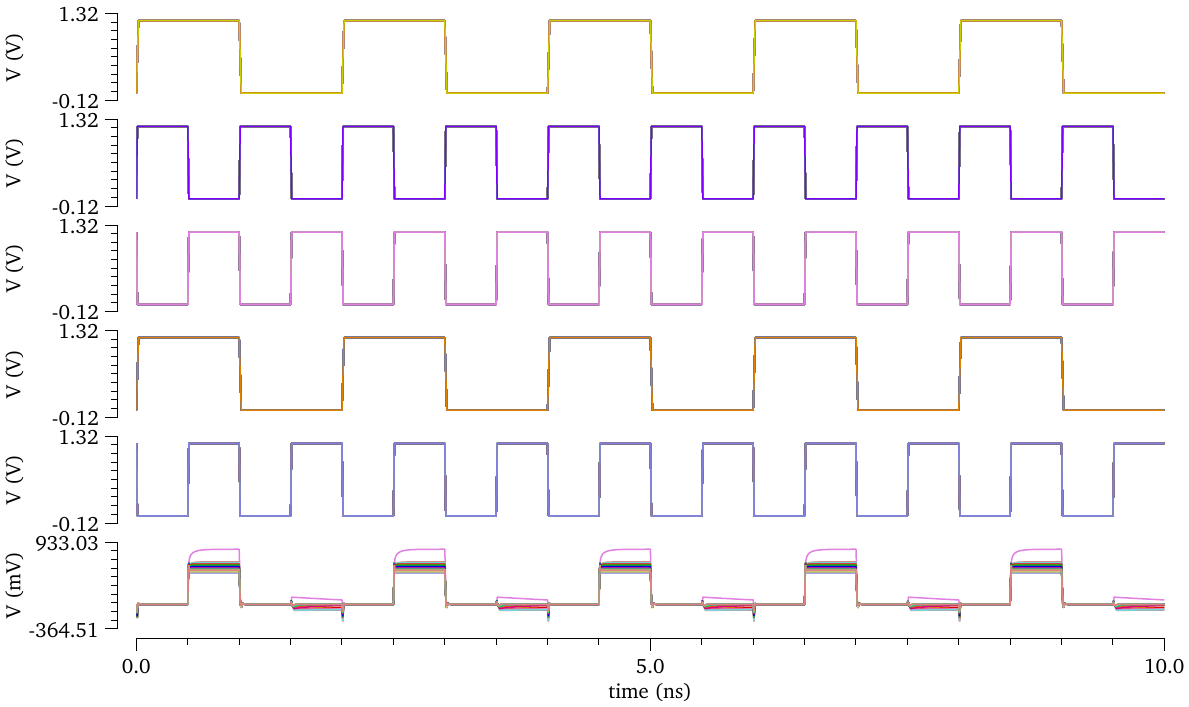}
    \label{hrscorner}}
    \centering
    \subfloat[]{\includegraphics[width=0.46\columnwidth, height=37.5mm]{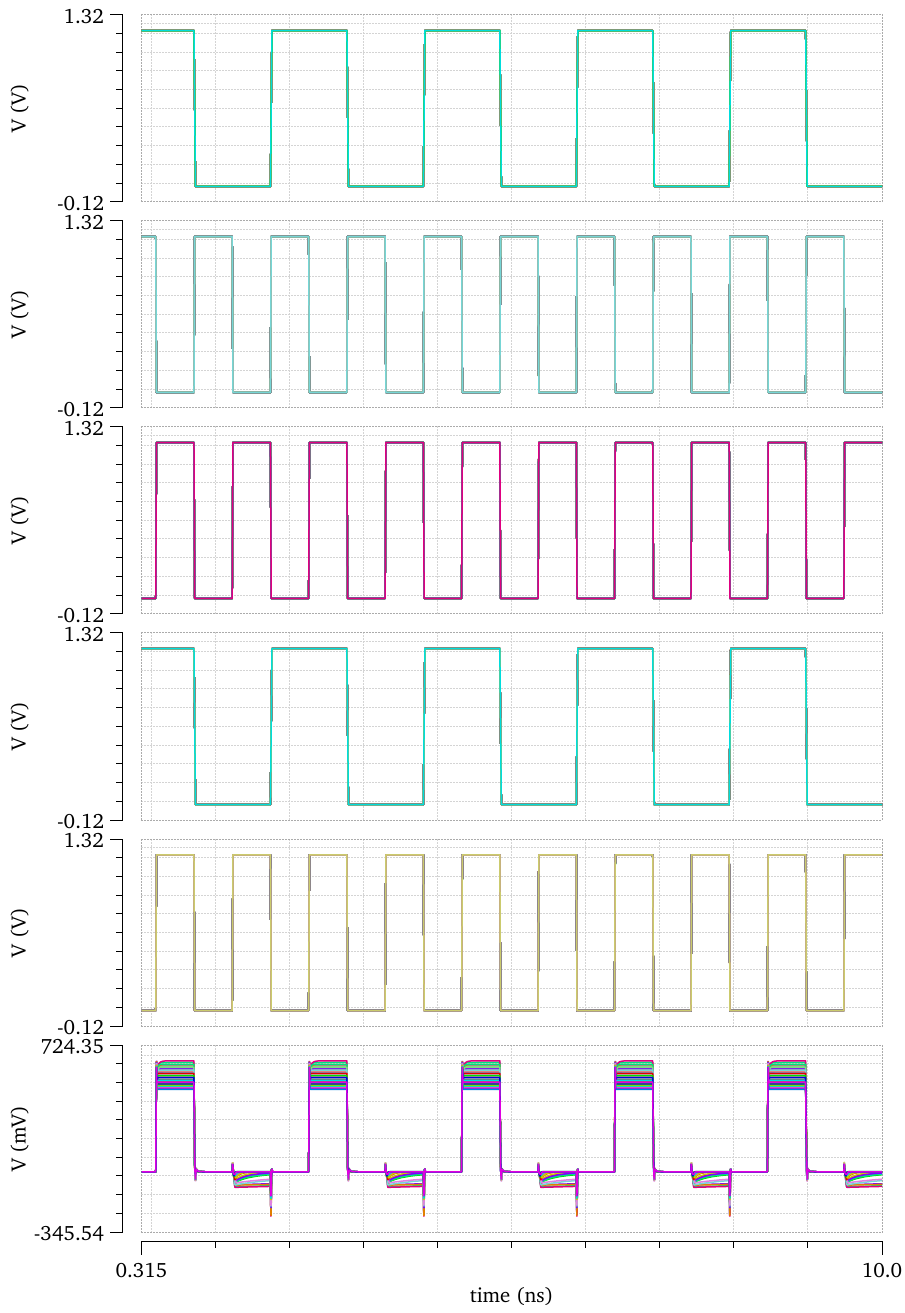}
    \label{lrscorner}}
    \hspace{0.1\columnwidth}
    \caption{Process, voltage, temperature, and ReRAM variability for the proposed 3T1R bitcell under TT, SS, FF, SF, and FS corners from -40$^{\circ}$C to 125$^{\circ}$C with $\pm$20\% HRS/LRS variation.}

\end{figure}

\begin{figure*}[!t]
\centering
\includegraphics[width=0.825\textwidth]{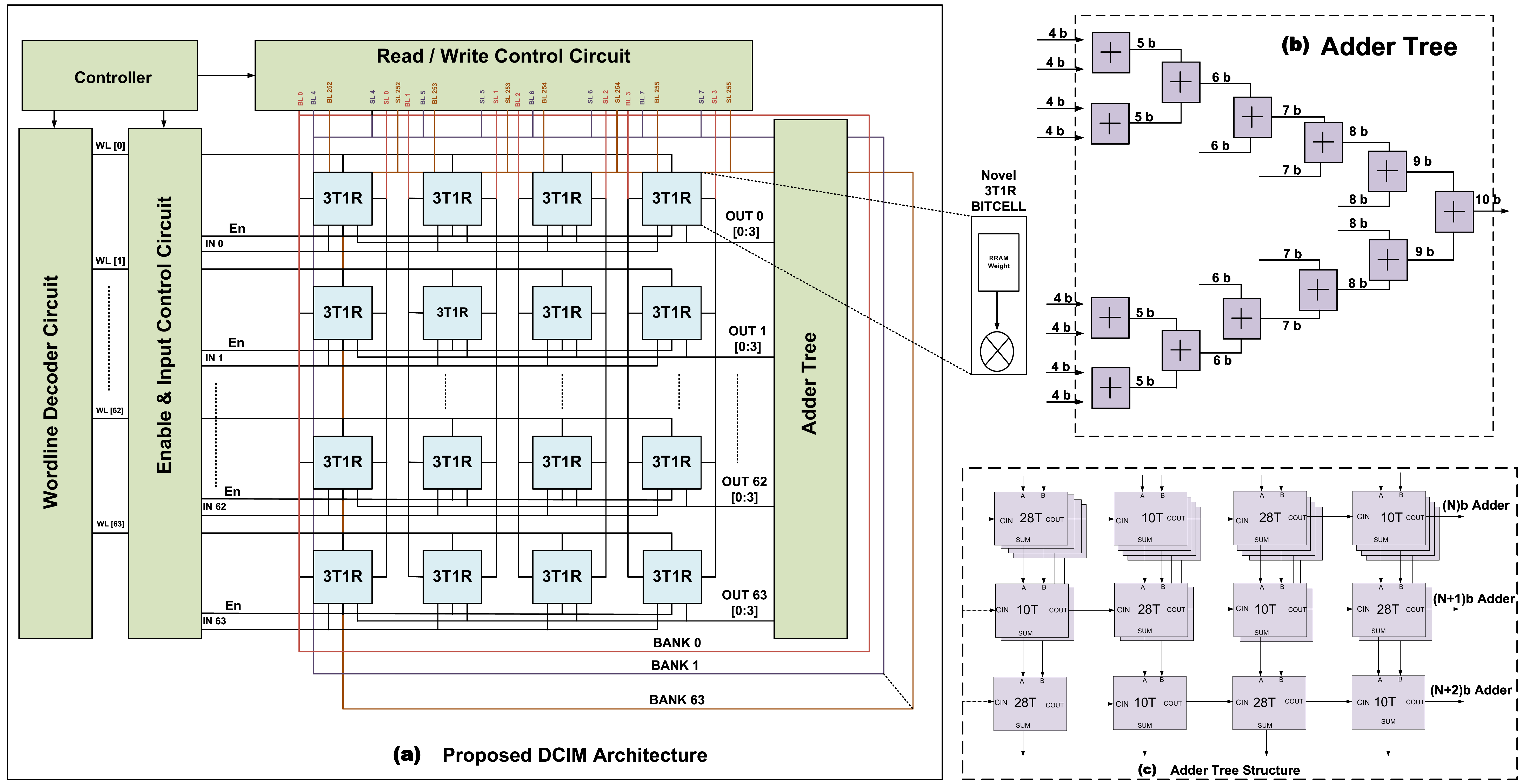}
\caption{(a) Proposed energy- and resource-efficient ReRAM-based DCIM macro, (b) accumulation flow of the interleaved adder tree, and (c) structure of the proposed 10T/28T interleaved adder tree.}

\label{Macro}
\end{figure*}

\section{Proposed Work}

\subsection{Novel ReRAM-based 3T1R Bitcell}

\textbf{Write, Read and CIM Operations:} The proposed ReRAM device operates between high-resistance state (HRS) and low-resistance state (LRS), controlled through SET and RESET voltages. A positive bias forms a conductive filament and switches the device to LRS, while a negative bias ruptures the filament and returns it to HRS. Unlike conventional iterative write-and-verify approaches, the proposed 3T1R nvCIM cell uses a large resistance window of nearly 50$\times$, enabling reliable digital sensing with a low-voltage read operation. In this work, the SET/RESET voltage is chosen as 1.2 V with LRS and HRS values of 10 k$\Omega$ and 500 k$\Omega$, respectively \cite{sharma202264, lu2020reram, chen2020reconfigurable}. The ReRAM model is implemented using Verilog-A and integrated with a 65 nm CMOS process based on prior modelling approaches \cite{tenwar-vdat, sharma202264}.

Fig. 1 illustrates the proposed 3T1R bitcell consisting of one selector transistor and two additional transistors for implementing local AND-based multiplication. Compared to multi-ReRAM-cell structures, the proposed design achieves lower area and write energy overhead while preventing crosstalk in large arrays \cite{bengel2020variability, chang20173t1r}. During CIM operation, the input activation drives transistor M2 to generate the multiplication output locally, as shown in Table \ref{tab:mac-compute}. During write operation, CIM\_EN remains disabled to reduce unnecessary switching power. The proposed 3T1R-AND bitcell occupies only 0.85 $\mu$m\textsuperscript{2}, while the 1×4 layout demonstrates scalability toward larger memory arrays. The simulated waveforms in Fig. \ref{fig:LRS} illustrate the behaviour of the bitcell during multiplication. Input logic ‘1’ is represented by V\textsubscript{BL}=low and V\textsubscript{SL}=high, whereas logic ‘0’ is represented by the opposite condition. When WL and CIM\_EN are enabled, the bitcell performs local AND-based multiplication between the stored weight and the input activation. The resulting node transitions confirm stable multiplication functionality across both HRS and LRS states.

\subsection{Process and ReRAM Variability Analysis}

A major concern in ReRAM-CIM design is variability arising from filament formation and rupture, which affects the HRS/LRS resistance values. To capture this effect, ±20\% variation was introduced into the ReRAM Verilog-A model for both HRS and LRS states. In addition, the proposed bitcell was evaluated across all CMOS process corners (TT, SS, FF, SF, FS) and temperatures from -40 to 125$^{\circ}$C. Despite these combined variations, the proposed bitcell maintains reliable digital functionality because of its large sensing margin and digital readout behaviour. As shown in Fig. \ref{lrscorner} and Fig. \ref{hrscorner}, the bitcell responds correctly under all process, voltage, temperature, and ReRAM variability conditions without requiring additional error-correction or reference-tuning circuitry.

\subsection{Interleaved 10T/28T Adder Tree}

The area overhead of the adder tree remains one of the major challenges in DCIM architectures. Conventional ripple-carry adders require 28 transistors per full adder, which increases both area and power. To address this issue, this work uses a compact 10T full adder from \cite{pratham-apccas} combined with conventional 28T full adders in an interleaved manner.

The interleaved adder tree reduces transistor count by nearly 37\% and lowers power consumption by 28\% compared to a conventional 28T RCA-based structure \cite{Flex-DCIM, sankhe-isqed24}. Since the 10T full adder introduces voltage degradation due to pass-transistor logic, however alternates 10T and 28T full adders across successive RCA stages to restore signal integrity. In the first stage, the RCA begins with a 28T full adder followed by a 10T full adder. In subsequent stages, the arrangement is alternated such that voltage degradation does not propagate across the adder tree.

\subsection{ReRAM-Based 16KB Digital CIM Macro}

The proposed 16 Kb ReRAM-based DCIM macro is organised into 64 independent banks, each consisting of 64 rows and 4 columns. Each bank supports 4-bit weight precision, while higher precision can be achieved by combining multiple banks. The architecture supports bit-serial input precision from 1 to 8 bits, allowing flexible mapping of convolutional and fully connected layers. During operation, input bits are applied sequentially from the most significant bit (MSB) to the least significant bit (LSB). In each cycle, the ReRAM array computes partial products, which are accumulated using the proposed low-area interleaved adder tree and stored in local 10-bit accumulation registers. This bit-serial approach enables support for arbitrary input precision with only N computation cycles for N-bit inputs.

The proposed architecture also reduces peripheral overhead by eliminating ADC/DAC blocks. As a result, power consumption is distributed mainly across the accumulator, activation functions, pooling, batch normalisation, and control logic. The removal of ADC/DAC circuitry significantly improves throughput and energy efficiency while maintaining fully digital operation. Beyond conventional CNN inference, the proposed macro can also support event-driven SNN execution. Since SNN activations are binary spike events, the proposed 3T1R-AND bitcell can directly perform spike-weight multiplication without additional conversion circuitry. This enables efficient accumulation of spike-domain partial sums while preserving the low-power and ADC-less characteristics of the proposed architecture.

\begin{figure}[!t]
    \centering
    \includegraphics[width=0.75\columnwidth]{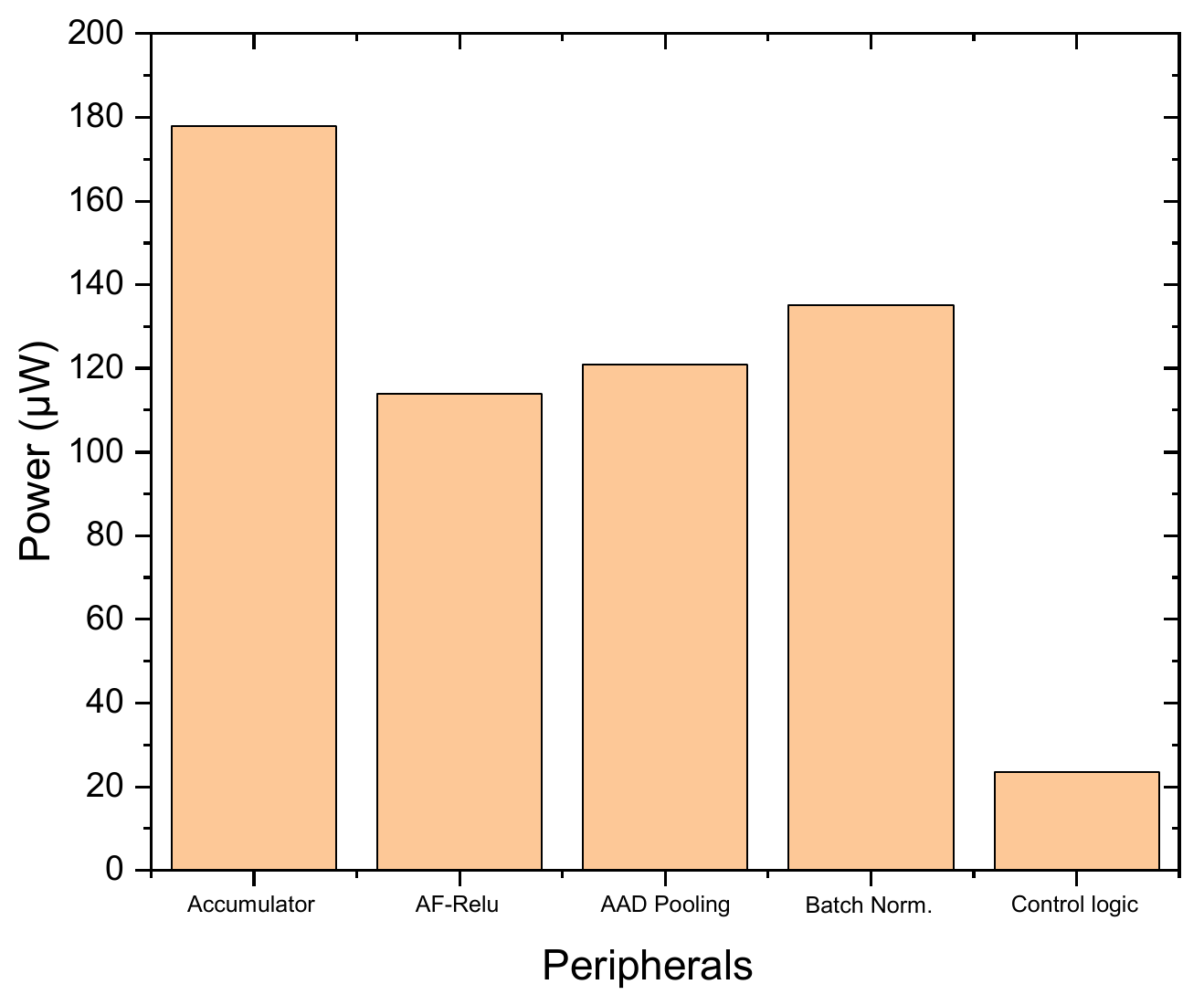}
    \caption{Distribution of peripheral power consumption across accumulation, activation, pooling, batch normalisation, and control circuitry in the proposed DCIM macro.}
    \label{fig:peripheral-power}
\end{figure}

\begin{table}[!t]
\caption{Comparison of different adder tree architectures in terms of average error, RMSE, power consumption, and delay in 65 nm CMOS at 1.2 V.}

\centering
\label{tab:adder-comp}
\resizebox{0.8\columnwidth}{!}{%
\begin{tabular}{l|c|c|c|c}
\hline
\textbf{Adder Tree Structure} & \textbf{Avg\_Error} & \textbf{RMSE} & \textbf{Power($\mu$W)} & \textbf{Delay (ns)} \\ \hline
\textbf{Conv-28T} & 0 & 0 & 892 & 3.56 \\ \hline
\textbf{Proposed} & 0 & 0 & 640 & 1.796 \\ \hline
\textbf{SESA\cite{SESA}} & 5.47 & 4.67 & 386.25 & 2.29 \\ \hline
\textbf{PG-26T\cite{Flex-DCIM}} & 0 & 0 & 586 & 2.5 \\ \hline
\textbf{2D-Inter-(7T+28T)\cite{sankhe-isqed24}} & 0.94 & 2.86 & 511 & 4.4 \\ \hline
\textbf{1D-Inter-(10T+28T)\cite{iscas-23}} & 1.34 & 3.1 & 673.89 & 7.34 \\ \hline
\textbf{Hybrid (28T + 16T)\cite{HOAA-VDAT24}} & 4.5 & 8.8 & 708 & 9.29 \\ \hline
\textbf{LOA (28T +OR)\cite{LOA-TVLSI}} & 1.38 & 4 & 488 & 4.7 \\ \hline
\end{tabular}}
\end{table}

\begin{table*}[!t]
\caption{Comparison of the proposed 16 Kb ReRAM-based DCIM macro with prior CIM architectures in terms of technology, bitcell type, latency, throughput, and energy efficiency.}
\label{tab:my-table}
\centering
\resizebox{\textwidth}{!}{%
\begin{tabular}{|c|c|c|c|c|c|c|c|c|c|c|}
\hline
\textbf{Parameters} & \textbf{TCAS-I'25\cite{Flex-DCIM}} & \textbf{VDAT'25\cite{tenwar-vdat}} & \textbf{ISQED'25\cite{sankhe-isqed24}} & \textbf{TCAS-I'22\cite{Vishal-tcas22}} & \textbf{ISCAS'24\cite{ISCAS'24}} & \textbf{TCAS-II'24\cite{VS-TCASII'24R}} & \textbf{TCAS-II'23\cite{3T2R-TCASII'23}} & \textbf{TCAS-II'21\cite{2T2R-CSL}} & \textbf{OJSCAS'21\cite{4T2R-OJCAS}} & \textbf{Proposed} \\ \hline
\textbf{Tech. (nm)} & 65 & 65 & 65 & 65 & 28 & 40 & 180 & 180 & 40 & 65 \\ \hline
\textbf{Supply Volt. (V)} & 1.0 & 0.7 & 1.2 & 0.8-1& 0.8–1.1 & 0.3 & 0.5–0.8 & 0.6–1.2 & 0.9 & 1.2 \\ \hline
\textbf{NVM Device} & SRAM & ReRAM & SRAM & ReRAM & ReRAM & ReRAM & ReRAM & ReRAM & ReRAM & ReRAM \\ \hline
\textbf{Operation} & AND & XNOR & AND & AND & MUL & MUL & MUL & MUL & CAM  & AND \\ \hline
\textbf{Bit-cell Area $\mu$m\textsuperscript{2} } & 2.34 & - & 4.1 & 0.847 & - & 0.085 & - & - & 0.55 & 0.85 \\ \hline
\textbf{Bit-cell Type} & 8T & 3T1R & 8T & 1T1R & 1T1R & 1T1R & 3T2R & 2T2R CSL & 4T2R & 3T1R \\ \hline
\textbf{Non-Volatility} & No & Yes & No & Yes & Yes & Yes & Yes & Yes & Yes & Yes \\ \hline
\textbf{Operation Mode} & Digital & Digital & Digital & Digital & Hybrid & - & Digital & Analog & Digital & Digital \\ \hline
\textbf{Precision} & 1-8b & 1-4b & 1-8b & 1-8/1-16 & - & - & 1b & 3b & - & 1-8b \\ \hline
\textbf{Latency (ns)} & 2.5 & 0.35 & 4.0 & 6 & 2.8 & 9.8 & 8.0 & 16.0 & 0.92 & 0.48 \\ \hline
\textbf{Throughput} & 0.4 & 2.86 & 2.2\textsuperscript{*} & - & 0.36 & 0.12 & 0.13 & 0.06 & 1.08 & 2.31, 3.1\textsuperscript{*} \\ \hline
\textbf{\begin{tabular}[c]{@{}c@{}}Energy-efficiency\\ (TOPS/W)\end{tabular}} & 249.1 & – & 480 & 114.4 & 5.82 & – & 289 & – & 223.6 & 314, 419 \\ \hline
\end{tabular}%
}
\end{table*}

\section{Methodology and Results Evaluation}

At the 65 nm CMOS technology node, the proposed 3T1R bitcell occupies an area of 0.85 $\mu$m\textsuperscript{2}. The proposed interleaved adder tree consumes an average power of 635 $\mu$W under a 1.2 V supply voltage.

To evaluate inference performance, the proposed macro was mapped to multiple neural networks, including LeNet-5 for MNIST and A-Z datasets, AlexNet for CIFAR-10, and CNN-8 for SVHN. The inference framework was implemented using QKeras in Python, where trained weights were exported in CSV format and mapped onto the proposed DCIM macro. Fig. \ref{accuracy} shows the inference accuracy under different precision settings, including 1A4W, 2A4W, 4A4W, and 8A8W. The results indicate only a minor accuracy degradation compared to the INT8 baseline. Among all precision settings, 2A4W provides the best tradeoff between computational efficiency and inference accuracy. In addition, pruning 40\% of the network parameters preserves nearly 99.8\% of the original accuracy while reducing MAC operations by 27\%, memory bank usage proportionally, and computation cycles by 33\%.

The hardware model incorporates array dimensions, precision configuration, bank mapping, and a layer-by-layer reuse strategy. The proposed DCIM macro processes one layer at a time, with weights programmed into the array dynamically. ReLU activation, pooling, softmax, and classification are handled by lightweight peripheral modules, enabling modular support across different CNN architectures. As an example, the second convolutional layer of LeNet-5 uses a 5$\times$5 kernel with 6 input channels and 16 output channels. Since each memory bank stores up to 64 weights, three banks are required for a complete filter set. Overall, Conv. 2 requires 32 banks and 48 computation cycles under 1A4W precision. Using this mapping strategy across all layers, the proposed architecture achieves up to 419 TOPS/W energy efficiency.

Further, from AND-based spike-domain CIM philosophy~\cite{spike_and}, the proposed macro was also evaluated for SNN-oriented workloads. Compared to the FP32 baseline, the proposed low-precision configurations exhibit only minor accuracy degradation. Since spike activations are binary in nature, the proposed AND-type bitcell can efficiently perform spike-weight multiplication with minimal switching activity. In particular, the 2A2W configuration provides accuracy very close to FP32 for all evaluated networks, including VGG-8 on CIFAR-10, VGG-16 on CIFAR-10, ResNet-18 on CIFAR-10, VGG-16 on ImageNet-1K, and ResNet-18 on CIFAR-100. The 4A4W and 8A8W configurations also maintain comparable performance with only marginal differences in Top-1 and Top-5 accuracy. Overall, the results indicate that moderate precision configurations provide an effective tradeoff between computational cost and inference accuracy, making them suitable for energy-efficient edge-AI deployment. The event-driven behaviour of SNN inference further reduces memory accesses, accumulator activity, and computation cycles, making the proposed macro highly suitable for low-power always-on sensing and neuromorphic edge applications. 

Table \ref{tab:my-table} shows that the proposed design improves latency and energy efficiency by nearly 30-40\% compared to prior state-of-the-art designs. The macro achieves a minimum latency of 0.48 ns, throughput of 2.31-3.1 TOPS, and energy efficiency of 314-419 TOPS/W with 30\% sparsity. In addition, the architecture supports configurable weight precision from 1 to 8 bits within a single 16 KB macro while eliminating the area and power overhead of ADC/DAC circuitry. Overall, the proposed E-ReCON macro demonstrates robust performance across PVT variations, supports flexible precision configurations, and maintains high inference accuracy with reduced memory and computational cost. These characteristics make it well suited for battery-powered edge-AI platforms.

\begin{figure}[!t]
\centering
\subfloat[]{\includegraphics[width=0.85\columnwidth]{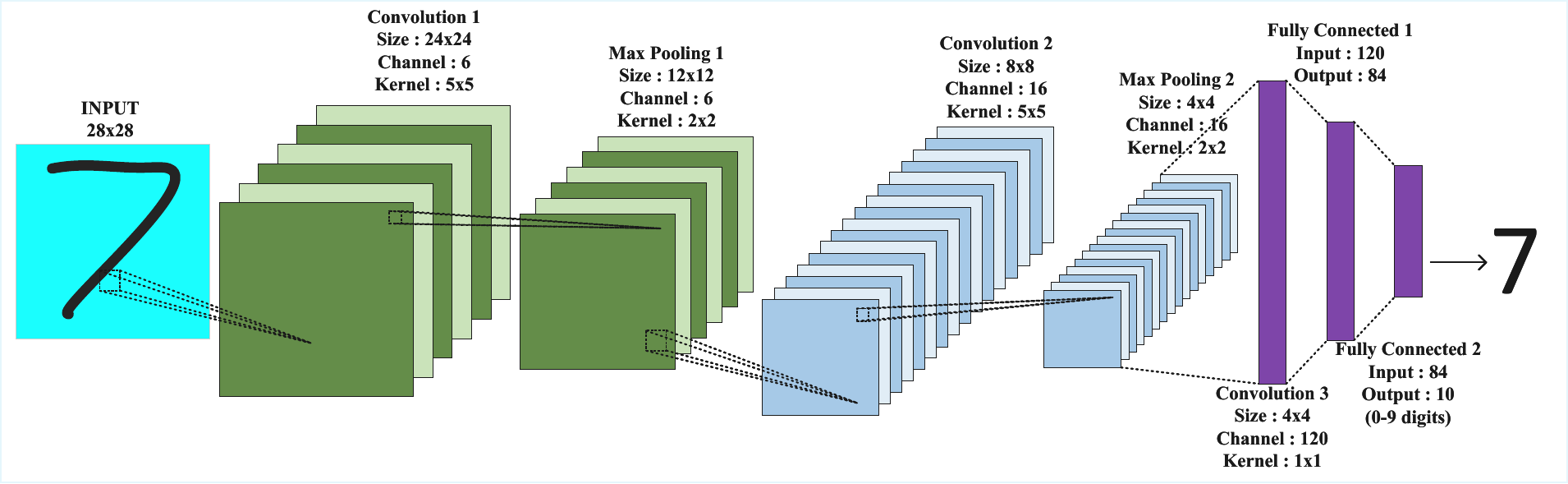}%
\label{lenet}}
\hfill
\subfloat[]{\includegraphics[width=0.85\columnwidth]{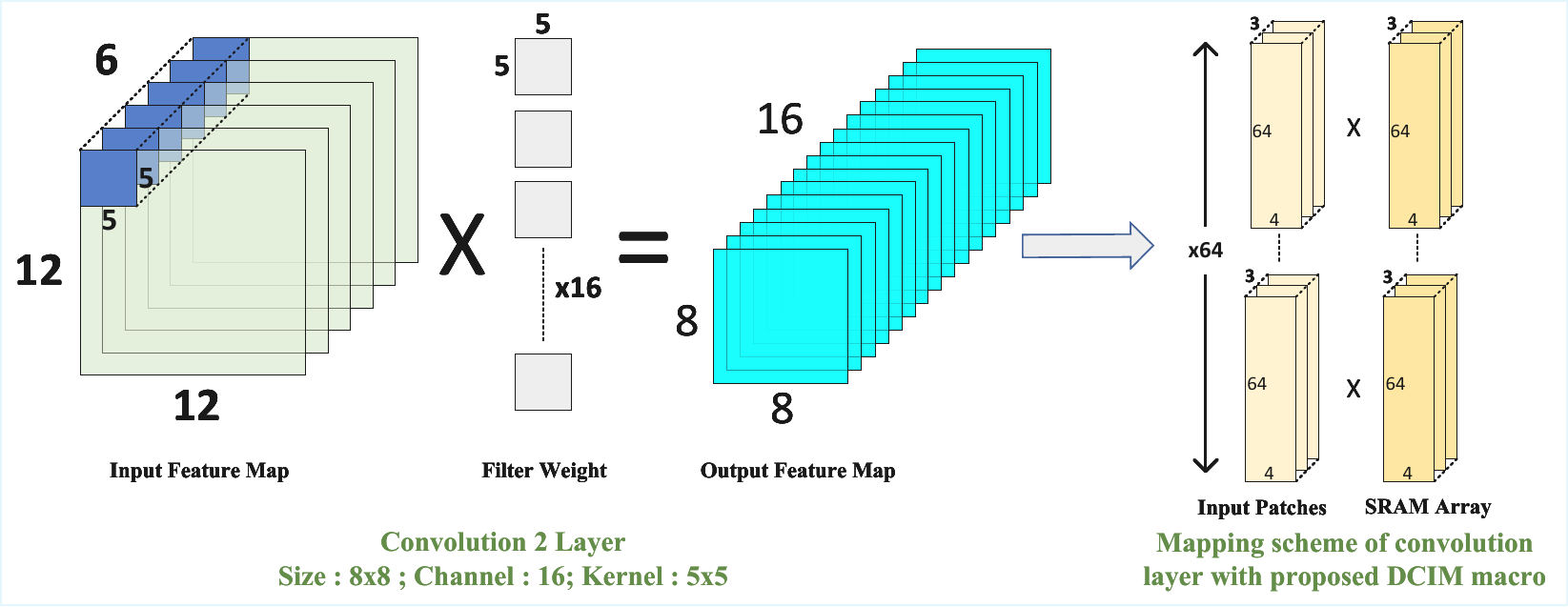}%
\label{mapping}}
\caption{(a) LeNet-5 network architecture used for evaluation, and (b) mapping of convolutional layers onto the proposed ReRAM-based DCIM macro.}
\end{figure}

\begin{table}[!t]
\caption{Layer-wise mapping of the pruned LeNet-5 network onto the proposed DCIM macro, including kernel size, output dimensions, bank allocation, and operation cycles.}
\label{mappingtable}
\renewcommand{\arraystretch}{1.2}
\scalebox{0.92}{
\begin{tabular}{|
>{\columncolor[HTML]{BFBFBF}}c |
>{\columncolor[HTML]{FFFFFF}}c |
>{\columncolor[HTML]{FFFFFF}}c |
>{\columncolor[HTML]{FFFFFF}}c |
>{\columncolor[HTML]{FFFFFF}}c |}
\hline
{\color[HTML]{000000} \textbf{Layer}} & \cellcolor[HTML]{BFBFBF}{\color[HTML]{000000} \textbf{Kernal}} & \cellcolor[HTML]{BFBFBF}{\color[HTML]{000000} \textbf{Output}} & \cellcolor[HTML]{BFBFBF}{\color[HTML]{000000} \textbf{Banks Used}} & \cellcolor[HTML]{BFBFBF}{\color[HTML]{000000} \textbf{Op Cycles}} \\ \hline
{\color[HTML]{000000} \textbf{Input}} & {\color[HTML]{000000} - \textbf{}} & {\color[HTML]{000000} $28 \times 28 \times 1$} & {\color[HTML]{000000} NA} & {\color[HTML]{000000} -} \\ \hline
{\color[HTML]{000000} \textbf{Conv. 1}} & {\color[HTML]{000000} $5 \times 5 \times 6$} & {\color[HTML]{000000} $24 \times 24 \times 6$} & {\color[HTML]{000000} $6$} & {\color[HTML]{000000} $384$} \\ \hline
{\color[HTML]{000000} \textbf{Max Pool 1}} & {\color[HTML]{000000} $2 \times 2$} & {\color[HTML]{000000} $12 \times 12 \times 6$} & \cellcolor[HTML]{FFFFFF}{\color[HTML]{000000} NA} & {\color[HTML]{000000} -} \\ \hline
{\color[HTML]{000000} \textbf{Conv. 2}} & {\color[HTML]{000000} $5 \times 5 \times 16$} & {\color[HTML]{000000} $8 \times 8 \times 16$} & {\color[HTML]{000000} $32$} & {\color[HTML]{000000} $48$} \\ \hline
{\color[HTML]{000000} \textbf{Max Pool. 2}} & {\color[HTML]{000000} $2 \times 2$} & {\color[HTML]{000000} $4 \times 4 \times 16$} & \cellcolor[HTML]{FFFFFF}{\color[HTML]{000000} NA} & {\color[HTML]{000000} -} \\ \hline
{\color[HTML]{000000} \textbf{Conv. 3}} & {\color[HTML]{000000} $1 \times 1 \times 120$} & {\color[HTML]{000000} $4 \times 4 \times 120$} & {\color[HTML]{000000} $45$} & {\color[HTML]{000000} $16$} \\ \hline
{\color[HTML]{000000} \textbf{Flatten}} & {\color[HTML]{000000} NA} & {\color[HTML]{000000} $1920 \times 1$} & \cellcolor[HTML]{FFFFFF}{\color[HTML]{000000} NA} & {\color[HTML]{000000} -} \\ \hline
{\color[HTML]{000000} \textbf{FC 1}} & {\color[HTML]{000000} $84$} & {\color[HTML]{000000} $10$} & {\color[HTML]{000000} $168$} & {\color[HTML]{000000} $3840$} \\ \hline
{\color[HTML]{000000} \textbf{FC 2}} & {\color[HTML]{000000} $10$} & {\color[HTML]{000000} $1$} & {\color[HTML]{000000} $20$} & {\color[HTML]{000000} $64$} \\ \hline
\end{tabular}
}
\end{table}

\begin{figure}[!t]
    \centering
    \subfloat[]{\includegraphics[width=0.48\columnwidth]{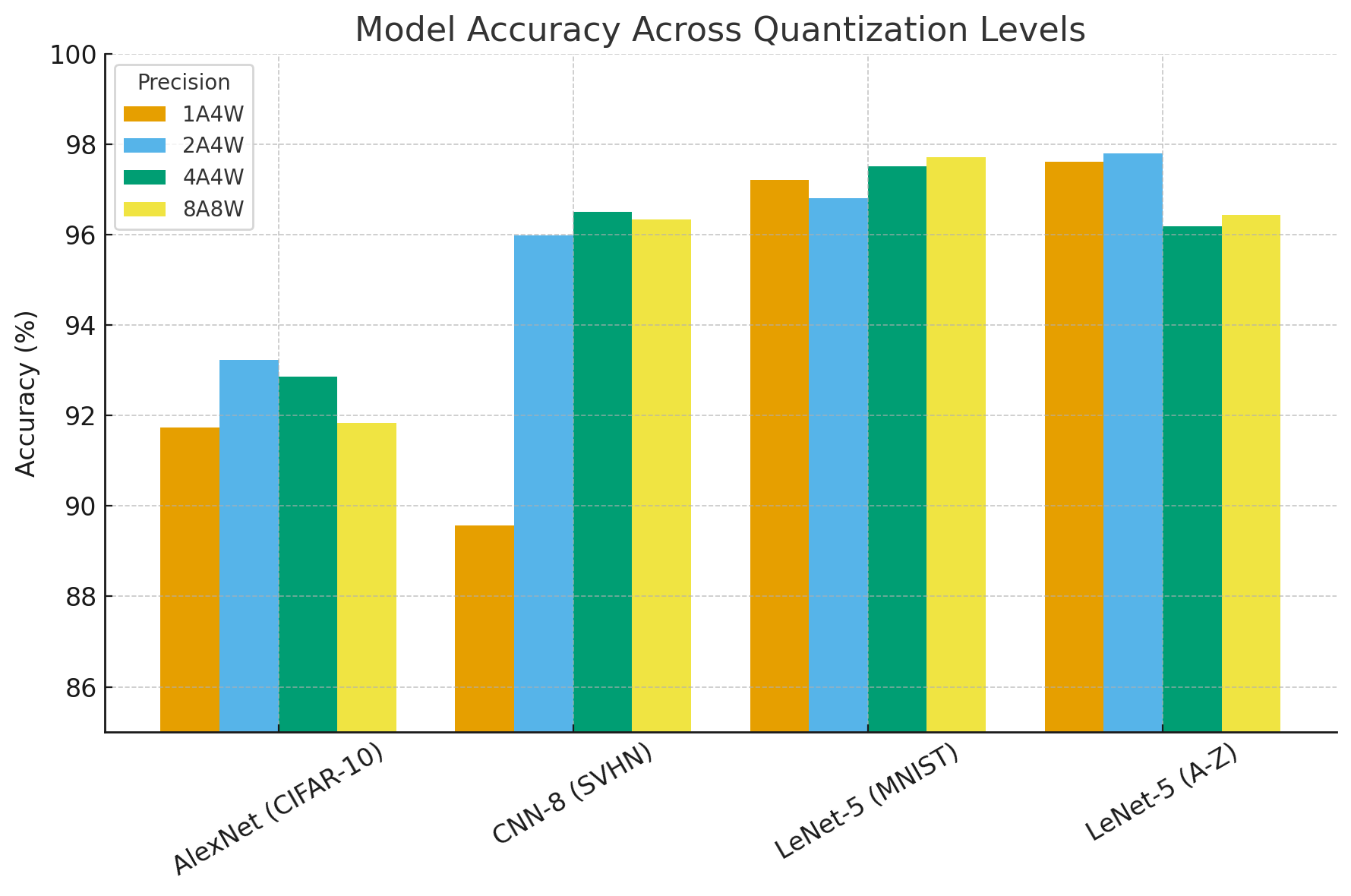}
        \label{accuracy}}
    \centering
    \subfloat[]{\includegraphics[width=0.44\columnwidth]{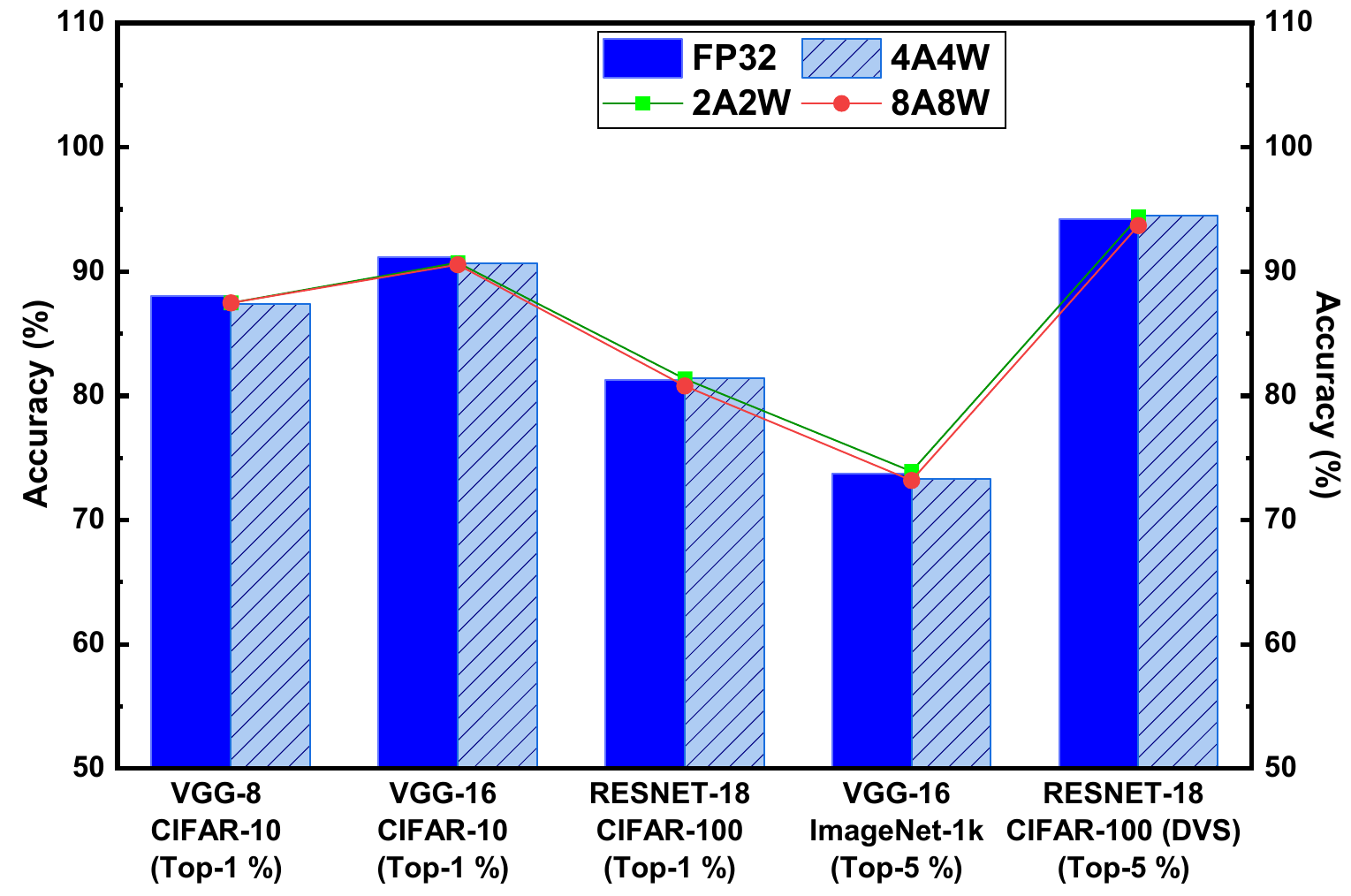}
    \label{fig:HRS}}
    \caption{(a) Accuracy comparison of CNN models across different quantisation settings for AlexNet on CIFAR-10, CNN-8 on SVHN, and LeNet-5 on MNIST and A-Z datasets, and (b) accuracy comparison of equivalent SNN models under different low-precision configurations.}
\end{figure}

\section{Conclusion}

This work presents E-ReCON, a compact and energy-efficient ReRAM-based DCIM macro for edge-AI inference. The proposed architecture employs a 3T1R AND-type ReRAM bitcell to eliminate ADC/DAC overhead while enabling reliable in-memory multiplication for both CNN and SNN workloads. In addition, a novel interleaved 10T/28T adder tree reduces transistor count and power consumption by 37\% and 28\%, respectively, compared to a conventional 28T RCA-based structure.

Implemented in 65 nm CMOS, the proposed macro achieves a minimum latency of 0.48 ns, throughput of 2.31-3.1 TOPS, and peak energy efficiency of 419 TOPS/W. The architecture achieves 97.81\% accuracy on LeNet-5, 93.23\% on AlexNet for CIFAR-10, and 96.51\% on CNN-8 for SVHN under low-precision configurations. In addition, 40\% pruning preserves nearly 99.8\% of the original accuracy while reducing MAC operations, memory usage, and computation cycles.

For SNN workloads, the proposed bitcell supports spike-domain multiplication and event-driven accumulation without additional conversion circuitry. The 2A2W precision setting achieves accuracy close to FP32 across VGG-8, VGG-16, and ResNet-18 models, while event-driven inference reduces memory accesses, switching activity, and accumulation overhead. Overall, E-ReCON demonstrates strong robustness against process and ReRAM variability while supporting 1-8 bit precision. These characteristics make it suitable for battery-powered edge-AI, IoT, and neuromorphic applications.

\bibliographystyle{ieeetr}
\bibliography{bib}

\end{document}